\documentclass{article}

\usepackage[nonatbib]{nips_2017}

\usepackage{graphicx} 
\usepackage{makecell}
\usepackage[utf8]{inputenc} 
\usepackage[T1]{fontenc}    
\usepackage{hyperref}       
\usepackage{url}            
\usepackage{booktabs}       
\usepackage{amsfonts}       
\usepackage{nicefrac}       
\usepackage{microtype}      
\usepackage{amsbsy}
\usepackage{amsmath}
\usepackage{pifont}
\usepackage{subcaption}
\usepackage{multirow}
\usepackage{tabu}
\usepackage[flushleft]{adjustbox}
\newcommand{\commenta}[1]

\title{Towards Interrogating Discriminative Machine Learning Models}

\vspace{-10mm}
\author{
  Wenbo Guo\\
  Pennsylvania State University\\
  \texttt{wzg13@ist.psu.edu} \\
  \And
  Kaixuan Zhang\\
  Pennsylvania State University\\
  \texttt{kuz22@ist.psu.edu} \\
  \AND
  Lin Lin\\
  Pennsylvania State University\\
  \texttt{llin@psu.edu} \\ 
  \And
  Sui Huang\\
  Netflix Inc.\\
  \texttt{shuang@netflix.com} \\
  \And
  Xinyu Xing\\
  Pennsylvania State University\\
  \texttt{xxing@ist.psu.edu} \\
}
  
\begin{document}
\maketitle
\vspace{-5mm}
\begin{abstract}
    It is oftentimes impossible to understand how machine learning models reach a decision. While recent research has proposed various technical approaches to provide some clues as to how a learning model makes individual decisions, they cannot provide users with ability to inspect a learning model as a complete entity. In this work, we propose a new technical approach that augments a Bayesian regression mixture model with multiple elastic nets. Using the enhanced mixture model, we extract explanations for a target model through global approximation.   To demonstrate the utility of our approach, we evaluate it on different learning models covering the tasks of text mining and image recognition. Our results indicate that the proposed approach not only outperforms the state-of-the-art technique in explaining individual decisions but also provides users with an ability to discover the vulnerabilities of a learning model.  
\end{abstract}
\vspace{-5mm}
\section{Introduction}
\label{sec:intro}
\vspace{-2mm}
Simpler machine learning (ML) methods\footnote{In this paper, we mainly consider discriminative machine learning but not other machine learning paradigms such as generative learning. Without further specification, by machine learning, we mean discriminative learning.} like decision tree and K-nearest neighbor have limited classification capability but provide better transparency~\cite{turner2016model}. As a result, they can provide end users with an explanation of individual decisions and even allow them to scrutinize model strengths and weaknesses. 

In comparison with those simple learning techniques above, complex learning models (e.g., deep neural networks) typically exhibit tremendous improvement in classification performance. However, they are almost completely opaque, even to the engineers that build them. Presumably as such, they have not yet been widely adopted in critical problem domains, such as diagnosing deadly diseases~\cite{Knight2017Dark} and making million-dollar trading decisions~\cite{Knight2017Financial}. 

To increase transparency for complicated learning models, recent research primarily adopts two kinds of mechanisms~\cite{gunning2017explainable} -- (1) {\em deep explanation} that alters learning models to produce more explainable representations, and (2) {\em model induction} that infers explanations for individual decisions through local  approximation. While both demonstrate a great potential to help users interpret an individual decision, they lack an ability to convey an understanding of how an ML model -- as a complete entity -- will behave in the future. As a result, they are not able to provide users with the ability to understand model strengths and weaknesses or, in other words, fail to enable users to foresee when prediction errors might occur. 


In this work, we propose a new technical approach. Different from the aforementioned two mechanisms, our approach not only explains an individual decision but, more importantly, provides a learning model with a comprehensive global explanation. As we will show in Section~\ref{sec:eval}, we can take the global explanation to craft adversarial (or pathological) examples and exploit model weaknesses. Technically, our approach introduces multiple elastic nets to a Bayesian regression mixture model and then uses it to extract explanations for models through global approximation. The intuition behind our approach is as follows. 

A Bayesian regression mixture model can approximate arbitrary probability density with high accuracy~\cite{marin2005bayesian}. As we will discuss in Section~\ref{sec:tech}, with multiple elastic nets, we can augment a regression mixture model with an ability to extract patterns even from a learning model that take as input high dimensional data with highly-correlated covariates. Given the pattern, we could extrapolate input features that are critical to the overall performance of an ML model. This information can be used to yield explanations for an individual decision and, more importantly, facilitate one to scrutinize a model's overall strengths and weaknesses. 

The rest of this paper is organized as follows. Section~\ref{sec:literature} surveys related work. Section~\ref{sec:tech} discusses our technical approach in detail. Section~\ref{sec:eval} describes our evaluation. We conclude this work in Section~\ref{sec:conclusion}. 


\commenta{

Different from {\tt LIME}~\cite{} -- the state-of-the-art approach that also follows the model induction mechanism -- our approach does not require sampling perturbed instances at random. Rather, it finds local explanation through a global approximation. As a result, it avoids approximation errors introduced by the sampling bias, making explanation more understandable to model creators and accountable to their users. We demonstrate this characteristic in Section~\ref{sec:eval}.

Going beyond providing explainability, we also introduce the elastic net~\cite{} to augment our regression mixture model with the ability to yield the patterns that a statistical model recognizes. Since the patterns identified uncover what parts of input features attribute the performance of a statistical model, one can use them to scrutinize model strengths and weaknesses. By taking advantage of patterns hidden behind a statistical model, Section~\ref{sec:eval} shows that, we can easily craft adversarial (or pathological) examples to exploit model weaknesses.

}
\vspace{-2mm}
\section{Related Work}
\label{sec:literature}
\vspace{-2mm}
As is discussed in Section~\ref{sec:intro}, prior endeavors in demystifying complicated ML models could be categorized into two types -- {\em deep explanation} and {\em model induction}. Here, we summarize these works and discuss their limitations respectively.

The deep explanation mechanism augments a learning model with the ability to yield explanations for individual predictions. Generally, the techniques in this kind mechanism follow two lines of approaches -- \ding{202} occluding portions of a single input sample and identifying what parts of the features are important for classification (e.g.,~\cite{li2016understanding, tamagnini2017interpreting, zeiler2014visualizing, zintgraf2017visualizing}), and \ding{203} computing a gradient of an output corresponding to a class with respect to a given input sample and pinpointing what features are sensitive to the prediction of that sample (e.g.,~\cite{gan2015devnet, selvarajugrad, simonyan2013deep, springenberg2014striving, sundararajan2016gradients}). While both can give users an explanation for a single decision that a learning model reach, they are not sufficient to provide a global understanding of a learning model, nor capable of exposing its strengths and weaknesses. Since most of the techniques following this mechanism requires altering a specific learning model, they typically cannot be generally applied to explaining prediction outcomes of other machine learning models. 

The model induction mechanism treats an ML model as a black box, and produces explanations by learning an interpretable model locally around a prediction. For example, Ribeiro et al. proposed {\tt LIME}~\cite{ribeiro2016should}, an explanation technique that samples perturbed instances around a single data sample and fits a sparse linear model to perform local explanations. Going beyond an explanation of a single prediction, Ribeiro et al. also attempted to extend their technique to explain a model as a complete entity. However, explanations obtained through this extension cannot describe the full mapping learned by a machine learning model in that {\tt LIME} explains a model as a whole only by selecting a small number of representative individual predictions and their explanations. 
\vspace{-2mm}
\section{Technical Approach}
\label{sec:tech}
\vspace{-2mm}
A Bayesian linear regression mixture model is a mixture of multiple Gaussian distributions
\begin{equation}
  \begin{aligned}
  \label{eq:GMM}
    y_i|\mathbf{x}_i, \boldsymbol{\Theta} \sim \sum_{j=1}^{\infty}\pi _{j}N(y_{i}\mid \mathbf{x}_{i}\cdot \boldsymbol{\beta}_{j},\sigma _{j}^{2}),
  \end{aligned}
\end{equation}
where  $\boldsymbol{\Theta}$ represents all parameters relevant and needed. $\mathbf{y} = (y_1,...,y_n)$ is a set of predictions for $n$ samples, and $\mathbf{X} = (\mathbf{x}_1, \mathbf{x}_2,...,\mathbf{x}_n) \in \mathcal{R}^{n\times p}$ is the corresponding sample feature matrix. Let  $\mathbf{x}_i = (x_{i1},..., x_{ip}) \in \mathcal{R}^{1 \times p}$ be the $i^{th}$ feature vector, and $\mathbf{X}_j$ be the $j^{th}$ column of $\mathbf{X}$ which contains the values of the $j^{th}$ feature across all the samples. $N(z|\mu_j, \tau_j)$ is the density of the univariate Gaussian distribution for $z$ with mean $\mu_j$, and variance $\tau_j$. $\pi_{1:\infty}$ are the component probabilities, with the sum equal to 1. $\boldsymbol{\beta}_{1:\infty}$ and $\sigma _{1:\infty}^{2}$ represent the parameters of regression models, with $\boldsymbol{\beta}_j \in \mathcal{R}^{p\times 1}$ and $\sigma _{j}^{2} \in \mathcal{R}$.

In general, a mixture model can approximate any learning model with high accuracy, and be viewed as a combination of multiple regression models. Given learning model $g:\mathcal{R}^p \rightarrow  \mathcal{R}$,  we can therefore approximate $g(\cdot)$ with a mixture model using $\{ \mathbf{X}, \mathbf{y} \}$, a set of data samples as well as their corresponding predictions obtained from model $g$, i.e., $\mathbf{y}  = g(\mathbf{X})$\footnote{For multi-class classification tasks, it should be noted that this work approximates each class separately, and therefore $\mathbf{X}$ denotes the samples in the same class and $g(\mathbf{X})$ represents the corresponding predictions. Note that if $\mathbf{y}$ is a probability vector, we should conduct logit transformation before fitting data into mixture model.}. For any data sample $\mathbf{x}_i$, we can then identify a regression model $y_i= \mathbf{x}_i \cdot \boldsymbol{\beta}_j +\epsilon$, which best approximates $g(\mathbf{x}_i)$, the prediction of $\mathbf{x}_i$. A more detail will be provided in the following.

Since regression models -- particularly linear regression -- have been used extensively for assessing how feature space affects a decision, by inspecting the weights (model coefficients) of the features present in the input, we can pinpoint the important features, and take them as an explanation for the corresponding individual decision. 

Going beyond the power of model approximation and local explainability, another characteristic of a mixture model is that, it can preserve only dominant patterns in data and enable multiple training data samples to share the same regression model. This significantly reduces the amount of explanations that a user has to inspect in order for him or her to scrutinize strengths and weaknesses of a model.

While a Bayesian regression mixture model provides us with a great potential to not only explain individual decisions but also understand a model as a whole, it does not always guarantee a success in model approximation, especially when data dimensionality is high, and feature space is sparse and highly correlated. For example, a conventional Bayesian regression mixture model cannot approximate a deep neural network that takes as input a sparse, high-dimensional image sample with highly correlated pixels (e.g., handwritten digits from MNIST~\cite{lecun1998gradient}).

To tackle this challenge, we introduce multiple elastic nets to a Bayesian regression mixture model. The elastic net is a regularized regression method that linearly combines the $l_{1}$ and $l_{2}$ penalties of the lasso and ridge methods. Past research~\cite{hans2011elastic, li2010bayesian} has demonstrated it can encourage the grouping effects among covariates so that highly correlated variables tend to be in or out of a mixture model together. As such, it can potentially augment the aforementioned model approximation method with the ability of dealing with the situation where a high dimensional data sample is sparse, and its features are highly correlated. In the following, we provide more details of a Bayesian regression mixture model. Then, we describe how we integrate multiple elastic nets into  this model, and discuss the novelty of our multiple elastic-net integration.

\subsection{Bayesian Non-parametric Regression Mixture Model}

As is specified in Equation~\eqref{eq:GMM}, the amount of Gaussian distributions is infinite. This indicates there are an infinite number of parameters that need to be estimated. In practice, the amount of data samples available is limited and therefore it is necessary to bound the number of distributions. To do this, truncated Dirichlet process prior \cite{ishwaran2001gibbs} can be applied, and the equation~\eqref{eq:GMM} can be expressed as follows
\begin{equation}
  \begin{aligned}
  \label{eq:Truncated_GMM}
    y_i|\mathbf{x}_i, \boldsymbol{\Theta} \sim \sum_{j=1}^{J}\pi _{j}N(y_{i}\mid \mathbf{x}_{i}\cdot \boldsymbol{\beta}_{j},\sigma _{j}^{2}).
  \end{aligned}
\end{equation}

To estimate parameters $\boldsymbol{\Theta}$ shown in the equation, a non-parametric Bayesian approach first models $\pi_{1:J}$ as random probabilities through a so-called ``stick-breaking'' prior process. With this modeling, parameters $\pi_{1:J}$ can then be computed by 
\commenta{
\begin{equation}
  \begin{aligned}
  \label{eq:stick_breaking}
    \pi_j = \left\{\begin{matrix}
         u_1&  & \text{for }& j = 1& \\ 
         u_j&\!\!\!\!\!\!\prod_{l = 1}^{j-1}(1-u_l) & \text{for }& j = 2&\!\!\!\!\!\!,...,J-1 \\
         1&\!\!\!\!\!\!\!\!\!\!\!\!-\sum_{l=1}^{J-1}\pi_l  & \text{for }& j = J&
            \end{matrix}\right.
  \end{aligned}
\end{equation}
}
\begin{equation}
  \begin{aligned}
  \label{eq:stick_breaking}
      \pi_j = u_j\prod_{l = 1}^{j-1}(1-u_l) \quad \text{for } j = 2,...,J-1,
  \end{aligned}
\end{equation}

with $\pi_1 = u_1$ and $\pi_J = 1-\sum_{l=1}^{J-1}\pi_l$. Here, $u_{k}$ follows a beta prior distribution, $\text{Beta}(1, \alpha)$ parameterized by $\alpha$, where $\alpha$ can be drawn from $\text{Gamma}(e,f)$, a Gamma conjugate prior with hyperparameters $e$ and $f$. To make computation efficient, $\sigma^{2}_j$ is set to follow an inverse Gamma prior, i.e., $ \sigma^2_{j} \sim \text{Inv-Gamma}(a,b)$, where $a$ and $b$ are hyperparameters. Given $\sigma^{2}_{1:J}$, for conventional Bayesian regression mixture model, parameters $\boldsymbol{\beta}_{1:J}$ can be drawn from Gaussian distribution $N(\mathbf{m}_{\beta},\sigma^{2}_{j}\mathbf{V}_{\beta})$ with hyperparameters $\mathbf{m}_{\beta}$ and $\mathbf{V}_{\beta}$.

As is described above, using a mixture model to approximate a learning model, for any data sample we can identify a regression model to best approximate the prediction of that sample. This is due to the fact that a mixture model can be interpreted as arising from a clustering procedure which depends on underlying latent indicators $z_{1:n}$. For each observation $(\mathbf{x}_{i}, y_{i})$, $z_i=j$ indicates that observation was generated from the $j^{th}$ Gaussian distribution. That is $y_i|z_i = j \sim N(\mathbf{x}_{i}\cdot \boldsymbol{\beta}_{j},\sigma _{j}^{2})$ with $Pr(z_{i}=j)=\pi_{j}$.

\subsection{Mixture Model with Multiple Elastic Nets}

Recall a conventional regression mixture model has difficulty in dealing with high dimensional data with highly correlated, sparse features. Here, we propose to enhance a mixture model with multiple elastic nets. Different from previous research~\cite{yang2011multiple} that also uses multiple elastic nets to handle high dimensional and highly correlated data, we design our approach to accommodate different types of data heterogeneity. In other words, instead of letting all data samples share the same elastic net parameters, we establish a new hierarchical Bayesian model that allows the parameters of multiple elastic nets to be categorized into multiple states of a Bayesian elastic net prior. As such, our approach has the flexibility to reduce a mixture model to Bayesian lasso or ridge regression under some sample categories, while maintaining the properties of the elastic net under other sample categories. In the following, we describe how we augment conventional regression mixture model with multiple elastic nets. 

We modify the conventional regression mixture model by resetting the prior distribution of $\boldsymbol{\beta}_{1:J}$ to realize multiple elastic nets. More specifically, we first define mixture distribution
\begin{equation}
  \begin{aligned}
  \label{eq:multiple elastic net}
  \pi(\boldsymbol{\beta}_{j} | \lambda_{1_{1:K}}, \lambda_{2_{1:K}}, \sigma^2_{j})=\sum_{k=1}^{K}w_{k}f_{k}(\boldsymbol{\beta}_{j}| \lambda_{1_{k}}, \lambda_{2_{k}}, \sigma^2_{j}),
  \end{aligned}
\end{equation}
where $K$ denotes the total number of component distributions, and $w_{1:K}$ represents component probabilities with $\sum_{k=1}^{K}w _{k}=1$. Let $w_{k}'s$ follow a Dirichlet distribution, i.e., $w_{1},w_{2},\cdots,w_{K} \sim \textrm{Dir}(1/K)$. $f_k$ is the Orthant Gaussian prior introduced in~\cite{hans2011elastic}, which can be  expressed  as follows
\begin{equation}
  \begin{aligned}
  \label{eq:multiple elastic net distribution}
  f_{k}(\boldsymbol{\beta}_{j}| \lambda_{1_{k}}, \lambda_{2_{k}}, \sigma^2_{j}) \propto \boldsymbol{\Phi} (\frac{-\lambda_{1_{k}}}{2\sigma\sqrt{\lambda_{2_{k}}}})^{-p}\times \sum_{\mathbf{Z} \in \mathcal{Z}}{N}(\boldsymbol{\beta}_{j}| -\frac{\lambda _{1_{k}}}{2\lambda_{2_{k}}}\mathbf{Z},\frac{\sigma^2_j}{\lambda_{2_{k}}}\mathbf{I}_p)\mathbf{1}(\boldsymbol{
    \beta}_{j}\in \mathcal{O}_{\mathbf{Z}}).
  \end{aligned}
\end{equation}
Here, $\lambda_{i_{k}}(i=1,2)$ is a pair of parameters which control lasso and ridge regression for the $k^{th}$ component, respectively. We set both to follow Gamma conjugate prior with $\lambda _{1_{k}} \sim \text{Gamma} (R, V/2)$  and $\lambda _{2_{k}} \sim \text{Gamma} (L, V/2)$. Note that $R,L,V$ are hyperparameters.  $\boldsymbol{\Phi}(\cdot)$ is the cdf of univariate standard Gaussian distribution, and $\mathcal{Z}=\{ -1,+1\} ^p$ is a collection of all possible p-vectors with elements $\pm 1$. Let $Z_l =1$ for $\beta_{jl} \geq 0$ and $Z_l =-1$ for $\beta_{jl} < 0$. Then, $\mathcal{O}_{\mathbf{Z}}\subset \mathbb{R}^p$ can be determined by vector $\mathbf{Z} \in \mathcal{Z}$, indicating the corresponding orthant. 

With the aforementioned definition for mixture distribution, we now derive the prior for $\boldsymbol{\beta}_{1:J}$. In particular, we introduce a set of latent indicators $c_{1:J}$. For each parameter $\beta_{j}$, $c_{j}=k$ indicates that parameter follows distribution $f_{k}(\cdot)$ with $Pr(c_j = k) = w_k$. Using this and following the process introduced in~\cite{hans2011elastic}, we can obtain the following
\begin{equation}
  \begin{aligned}
  \label{eq:Prior7}
    \boldsymbol{\beta}_{j} \mid \boldsymbol{\tau}_{j},\sigma_{j}^{2},\lambda_{2_{c_{j}}} \sim {N}(\boldsymbol{\beta}_j \mid 0,\frac{\sigma_{j}^{2}}{\lambda_{2_{c_{j}}}}\mathbf{S}_{\boldsymbol{\tau}_{j}}),
  \end{aligned}
\end{equation}
\begin{equation}
  \begin{aligned}
  \label{eq:Prior8}
    \boldsymbol{\tau}_{j} \mid \sigma_{j}^{2},\lambda_{1_{c_{j}}},\lambda_{2_{c_{j}}} \sim \prod_{l=1}^{p}\text{Inv-Gamma}_{(0,1)}(\tau_{jl} \mid \frac{1}{2},\frac{1}{2}(\frac{\lambda_{1_{c_{j}}}}{2\sigma_{j}\sqrt{\lambda_{2_{c_{j}}}}})^{2}),
  \end{aligned}
\end{equation}
where $\boldsymbol{\tau}_{j} \in \mathbb{R}^{p \times 1}$ denotes latent variables and $\mathbf{S}_{\boldsymbol{\tau}_{j}}\in \mathbb{R}^{p\times p}$, $\mathbf{S}_{\boldsymbol{\tau}_{j}}=\textrm{diag}(1-\tau_{jl})$, $l=1,\cdots,p$.

\subsection{Posterior Computation and Post-MCMC Analysis}

We develop a customized MCMC method involving a combination of Gibbs sampling and Metropolis-Hastings for posterior computation. Briefly, it involves augmentation of the model parameter space by the aforementioned mixture component indicators, $z_{i}, i = 1,...,n$ and $c_{j}, j = 1,...,J$. These indicators enable simulation of relevant conditional distributions for model parameters. As the MCMC proceeds, they can be estimated from relevant conditional posteriors and thus we can obtain posterior simulations for model parameters and mixture component indicators jointly. Due to the page limits, we provide technical details in supplement material. Considering fitting a mixture model with MCMC suffers from the well-known label switching problem, complicating posterior inference. we use an iterative relabeling algorithm introduced in~\cite{cron2011efficient}.
\vspace{-2mm}
\section{Evaluation}
\label{sec:eval}
\vspace{-2mm}

Recall the motivation of our proposed solution is to increase transparency for complicated machine learning models so that users could leverage our solution to not only understand an individual decision (explainability) but also scrutinize the strengths and weaknesses of the target model (scrutability). Our evaluation of the proposed solution thus focuses on aforementioned two aspects -- explainability and scrutability.

\subsection{Experimental Setup}

We evaluate our proposed solution in the context of image recognition and text classification. More specifically, we apply our approach to explain deep neural network (DNN) for classifying handwritten digits, and random forest as well as Support Vector Machine (SVM) for categorizing newsgroup posts. These machine learning methods represent the techniques most commonly used for the corresponding classification tasks. Note that before we set out to understand aforementioned ML models using our approach, we first train these ML models to achieve more than decent classification performance on the training data sets. The following section describes the data sets that we used to train those classifiers as well as tactics and primary metrics that we use to evaluate the validity and utility of our proposed approach. 

\subsubsection{Data Set}

\noindent{\textbf{`comp.sys' newsgroups data set}~\cite{Lang95}:} 
It is a collection of newsgroups posts containing 1,945 samples across 2 topics. The newsgroups posts are split into training and testing data sets based on the dates they have been posted. It is a subset of 20 newsgroups data set~\cite{Lang95}.    


\noindent{\textbf{MNIST data set}~\cite{lecun1998mnist}:} 
This data set is a large database of handwritten digits that is commonly used for training various image recognition systems. It is composed of 70,000 greyscale images (of 28$\times$28, or 784, pixels) of handwritten digits, split into a training set of 60,000 samples and a testing set of 10,000 samples. 

\subsubsection{Tactics and Metrics}

Intuition suggests that close approximation is essential to precise insights into the model under examination. Therefore we develop our solution as a classifier treating target model predictions as ground truth and approximate the decision boundary. We measure Root Mean Square Error (RMSE) to gauge our approximation accuracy: $\sqrt{\frac{\sum_{i=1}^{n}(g_i-\hat{g_i})}{n}}$, where $g_i$ represents a single prediction obtained from a learning model, and $\hat{g_i}$ denotes the approximated prediction obtained from our approach. Here, $n$ is the total number of training data samples. To be specific, we calculate the RMSE from the-state-of-the-art solution in this field (i.e. {\tt LIME}) as the benchmark, against which we evaluate the superiority of our solution in terms of approximation accuracy. To also establish the faithfulness of our inferences to the target model, we apply our solution to machine learning models that are self explainable (i.e. Random Forest and SVM) and measure the consistency in feature importance inferences. To showcase how our proposed solution could enhance the scrutability of a target ML model, we manually craft adversarial and pathological examples against the patterns that our approach extracts and examine if they can exploit the weaknesses of corresponding learning models. The intuition behind this is that the exploitability of adversarial (and pathological) samples indicates the correctness of the patterns that our approach extracts. In the following, we describe our findings as well as the design of our experiments in greater details.

\begin{figure}[t]
    \centering
    \begin{subfigure}[t]{0.8\textwidth}
        \includegraphics[width=1\textwidth]{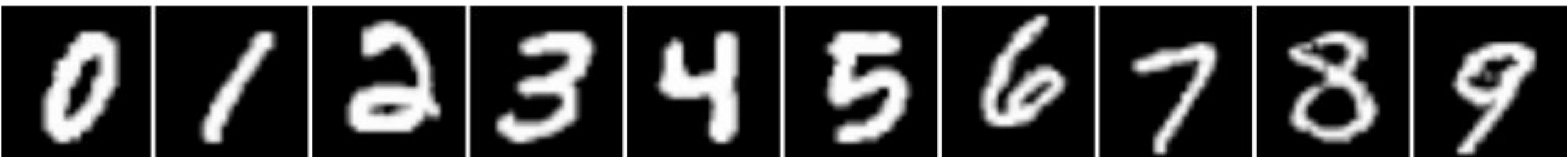}
        \caption{Handwritten digits randomly selected from MNIST data set.}
        \label{subfig:original}
    \end{subfigure} \\
    \begin{subfigure}[t]{0.8\textwidth}
        \includegraphics[width=1\textwidth]{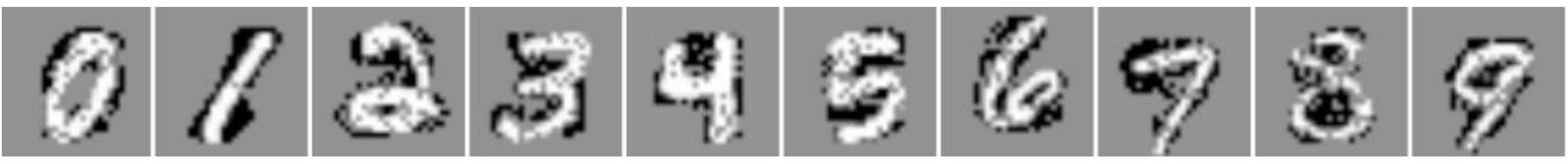}
        \caption{Most influential pixels highlighted by proposed approach.}
        \label{subfig:our}
    \end{subfigure} \\
    \begin{subfigure}[t]{0.8\textwidth}
        \includegraphics[width=1\textwidth]{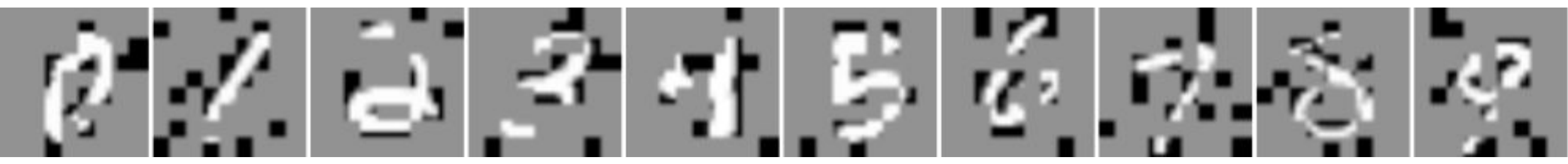}
        \caption{Most influential pixels highlighted by {\tt LIME}.}
        \label{subfig:lime}
    \end{subfigure}
    \caption{The examples explaining individual predictions obtained from a deep neural network trained for classifying handwritten digits. It should note that, to better illustrate the difference, we change pixels in grey if they are not selected.}
    \label{fig:feature_image}
\end{figure}

\begin{figure}[t]
\begin{center}
  \includegraphics[scale=0.33]{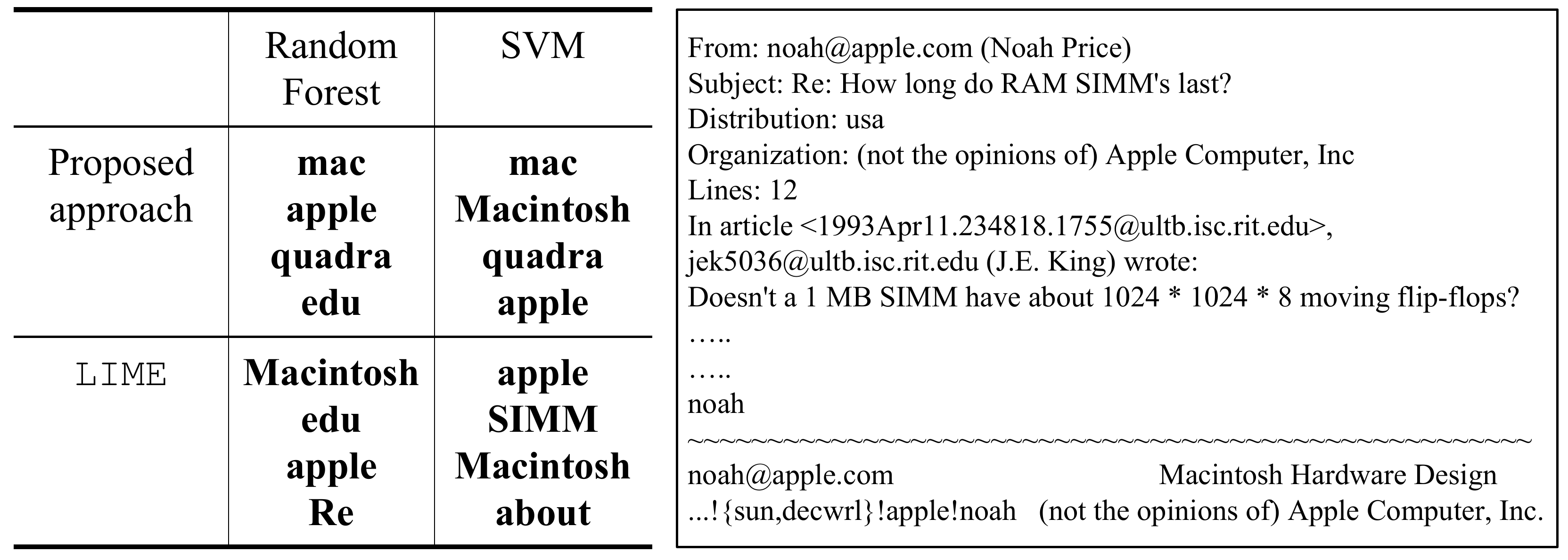}
\end{center}
\caption{The examples explaining individual predictions obtained from random forest and SVM trained for classifying `mac.hardware' news posts. Note that the text in bold indicate top-4 keywords most influential upon text classification.}
\vspace{-6mm}
\label{fig:feature_text}
\end{figure}

\begin{table}
\begin{subtable}[t]{0.9\textwidth}
\centering
\scriptsize
\begin{tabular}{ccccccccccc}
\Xhline{1.2pt}
{\begin{tabular}[c]{@{}c@{}} Technology\end{tabular}} & \multicolumn{10}{c}{Handwritten digit}                                                              \\ \cline{2-11} 
                                                                                & 0      & 1      & 2      & 3      & 4      & 5      & 6      & 7      & 8      & 9      \\ \hline
Proposed approach                                                                 & 0.0040 & 0.0264 & 0.0017 & 0.0033 & 0.0048 & 0.0022 & 0.0053 & 0.0118 & 0.0046 & 0.0045 \\ \hline
{\tt LIME}                                                                            & 0.4284 & 0.3151 & 0.2170 & 0.2270 & 0.2877 & 0.1530 & 0.3827 & 0.1969 & 0.2705 & 0.2324 \\ \Xhline{1.2pt}
\end{tabular}
\caption{DNN trained on image data set}
\label{subtab:error_image}
\end{subtable}
\\
\begin{subtable}[t]{0.43\textwidth}
\centering
\scriptsize
\begin{tabular}{ccc}
\Xhline{1.2pt}
\multirow{2}{*}{\begin{tabular}[c]{@{}c@{}}  Technology\end{tabular}} & \multicolumn{2}{c}{News topics of `comp.sys'}                              \\ \cline{2-3} 
                                                                               & \multicolumn{1}{l}{`ibm.pc.hardware'} & \multicolumn{1}{l}{`mac.hardware'} \\ \hline
Proposed approach                                                                & 0.2449                                & 0.1839                             \\ \hline
{\tt LIME}                                                                   & 0.4803                                & 0.3249                             \\ \Xhline{1.2pt}
\end{tabular}
\caption{Random forest trained on text data set}
\label{subtab:error_rf}
\end{subtable}
\qquad
\begin{subtable}[t]{0.43\textwidth}
\centering
\scriptsize
\begin{tabular}{ccc}
\Xhline{1.2pt}
\multirow{2}{*}{\begin{tabular}[c]{@{}c@{}} Technology\end{tabular}} & \multicolumn{2}{c}{News topics of `comp.sys'}                                      \\ \cline{2-3} 
                                                                               & \multicolumn{1}{l}{`ibm.pc.hardware'} & \multicolumn{1}{l}{`mac.hardware'} \\ \hline
Proposed approach                                                                & 0.1597                                    & 0.1172                                 \\ \hline
{\tt LIME}                                                                   & 0.2344                                    & 0.1221                                 \\ \Xhline{1.2pt}
\end{tabular}
\caption{SVM trained on text data set}
\label{subtab:error_svm}
\end{subtable}
\caption{The comparison of approximation errors across each class.}
\label{tab:error}
\end{table}

\subsection{Explainability}

Before we showcase how our proposed solution could complement complex ML models by providing explainability, we present the approximation accuracy (as measured by RMSE) of our solution in the aforementioned two scenarios (handwritten digit recognition and text classification) along with the corresponding performance from {\tt LIME} as our benchmark. As is shown in Table~\ref{tab:error}, our technical approach exhibits much lower RMSE than {\tt LIME}, which indicates superior approximation accuracy. The reason behind this observation is that, different from {\tt LIME}, our approach does not require random sample perturbations. As such, it avoids approximation errors introduced by the sampling bias. 

Better approximation brings about more precise and granular insights into the target model. Figure~\ref{subfig:original} illustrates ten handwritten digits randomly selected from each of the classes in MNIST data set. We apply our solution as well as {\tt LIME} to each of the images shown in the figure then select and highlight the top 150 influential pixels that each approach deems important to the decision made by deep neural network classifier. The results are presented in Figure~\ref{subfig:our} and~\ref{subfig:lime} for our approach and {\tt LIME} respectively.

As we can observe in these figures, our approach nearly perfectly highlights the contour of each digit, whereas {\tt LIME} identifies only the partial contour of each digit. We repeat this exercise for 100 times and find this superior performance persist through trials. This indicates our approach offers better resolution and more granular explanations to individual predictions. 

Similar to what we observe in handwritten figure recognition case, our approach also outperforms {\tt LIME} in the context of text classification. Figure~\ref{fig:feature_text} illustrates one such example: the words highlighted are the most influential indicators for determining if the text snippet belongs to the category of ``mac hardware''. By applying both our approach and {\tt LIME} to the random forest and SVM classifiers, we can observe that the keywords highlighted by our approach is intuitively more distinguishable than those identified by {\tt LIME}.

Since SVM and random forest models are explainable by nature, we also leverage this case to quantify our solution's faithfulness to a target model. To be specific, we identify top four influential words for each of the classification task using our solution and compare them with the top four most weighted features in the original model. While the order of top four influential words vary slightly in our comparison against SVM, the words that our solution identifies match perfectly with those revealed in SVM and random forest model.

\begin{figure}[t]
\begin{center}
  \includegraphics[scale=0.4]{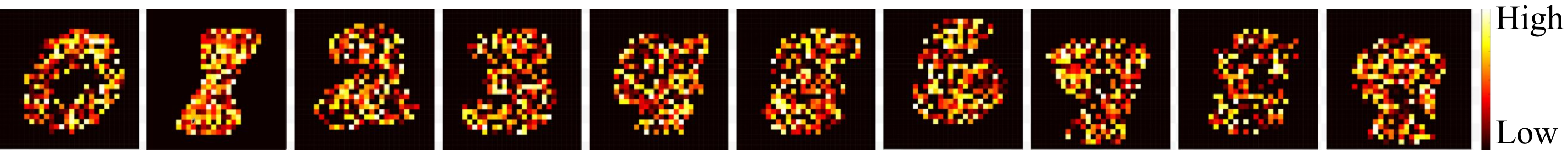}
\end{center}
\caption{The illustration of patterns extracted from the deep neural network trained for recognizing handwritten digits. Each pattern contains 150 pixels, the importance of which is illustrated in the form of a heat map. }
\label{fig:rule_img}
\end{figure}

\begin{figure}[t]
    \centering
    \begin{subfigure}[t]{0.48\textwidth}
        \includegraphics[width=1\textwidth]{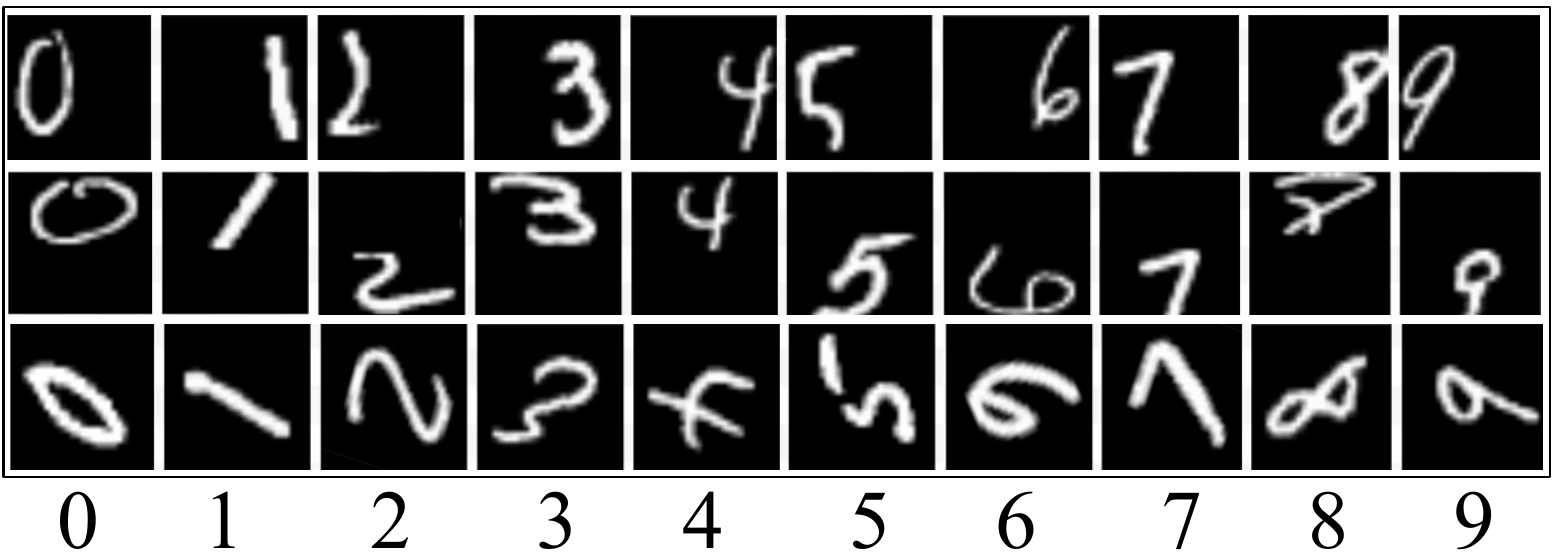}
        \caption{Adversarial samples.}
        \label{subfig:adv_image}
    \end{subfigure}
    \begin{subfigure}[t]{0.48\textwidth}
        \includegraphics[width=1\textwidth]{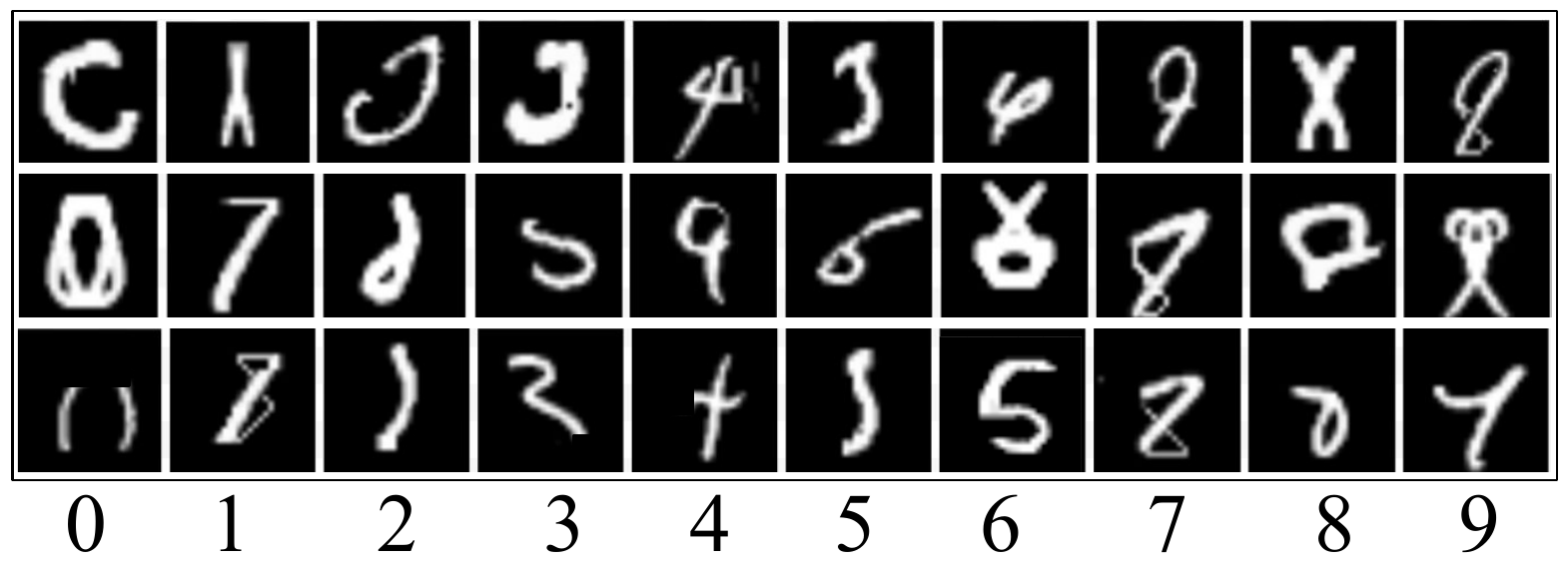}
        \caption{Pathological samples.}
        \label{subfig:path_image}
    \end{subfigure}
    \caption{Adversarial and pathological samples crafted based on the patterns illustrated in Figure~\ref{fig:rule_img}.}
    \vspace{-5 mm}
    \label{fig:adv_img}
\end{figure}

\begin{table}[t]
\centering
\begin{tabular}{cc|cc}
\Xhline{1.2pt}
\multicolumn{2}{c|}{Random Forest}                                          & \multicolumn{2}{c}{SVM}                                             \\ \hline
\multicolumn{1}{l}{`ibm.pc.hardware'} & \multicolumn{1}{l|}{`mac.hardware'} & \multicolumn{1}{l}{`ibm.pc.hardware'} & \multicolumn{1}{l}{`mac.hardware'} \\ \hline
`dos'                                 & `mac'                               & `ide'                                 & `mac'                              \\
`controller'                          & `apple'                             & `gateway'                             & `Macintosh'                        \\
`pc'                                  & `quadra'                            & `dos'                                 & `quadra'                           \\
`windows'                             & `edu'                               & `pc'                                  & `apple'                            \\ \Xhline{1.2pt}
\end{tabular}
\caption{The keywords that our approach extracts, indicating the features most imfulential upon classifications.}
\vspace{-6 mm}
\label{tab:rule_text}
\end{table}

\begin{figure}[t]
\begin{center}
  \includegraphics[scale=0.45]{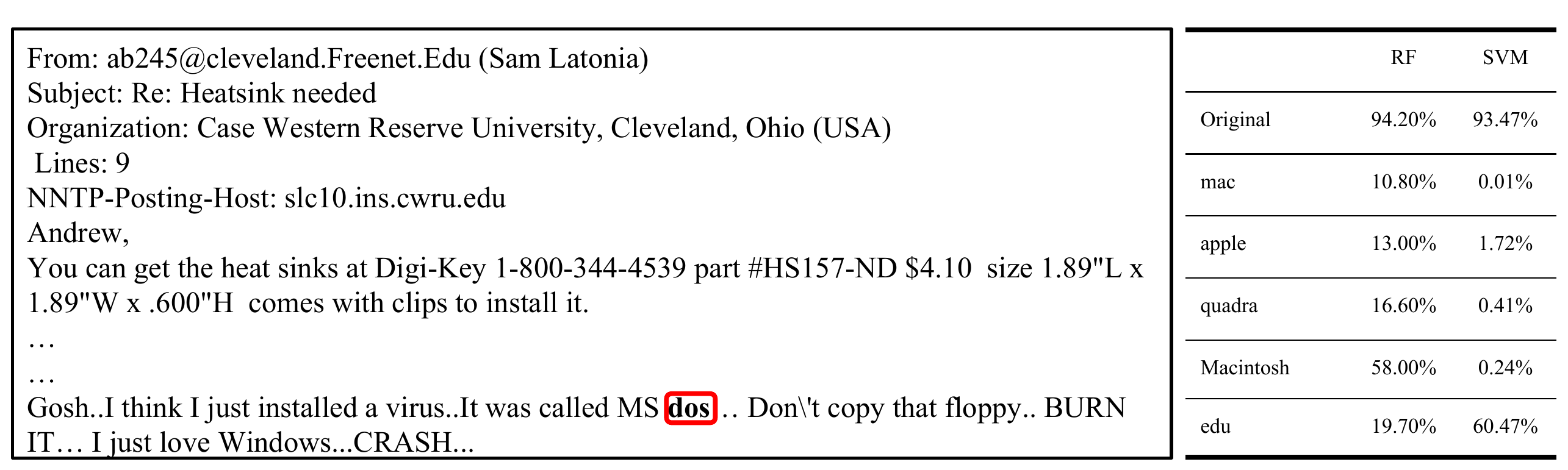}
\end{center}
\caption{An example text snippet categorized into `ibm.pc.hardware' by random forest (RF) and SVM (SVM). The percentages shown in the table indicate the confidence of being categorized in `ibm.pc.hardware'.}
\label{fig:adv_text}
\vspace{-6 mm}
\end{figure}

\subsection{Scrutability}

We define scrutability as an ability to not only identify inputs that are critical to the target ML model but also recognize dimensions that are relatively more vulnerable, or could lead to misclassification in a given ML model. To establish the scrutability of our proposed solution, we construct adversarial and pathological samples based on the vulnerable dimensions that our approach identifies and test them on the target ML model. 

Figure~\ref{fig:rule_img} illustrates patterns that our solution deems important to the aforementioned deep neural network model classifying handwritten digits in MNIST dataset in heat map form. While most patterns follow our human intuition, we notice that all the recognized patterns are distributed at the center of the canvas, which implies the model may only identify the digits written at the center. For some patterns (e.g., the pattern for digit one), we also observe the model may tolerate a handwritten digit with a certain degree of rotation. In addition, we discover that patterns extracted for some classes could be visually similar (e.g., digits 4 and 9), which might indicate that the model might have weaker distinguishability for some classes.

To verify our hypothesis above, we craft different adversarial samples, which are presented in  Figure~\ref{fig:adv_img}. To be specific, the first and second rows of adversarial samples are meant to test if this DNN model can only recognize digits that are located at the center of the canvas while the last row of adversarial samples are testing if the model only tolerates certain degree of rotation. These adversarial samples presented in Figure~\ref{fig:adv_img} are all misclassified by DNN with close to $99\%$ confidence.

In Figure~\ref{fig:adv_img}, we also illustrate some pathological samples that follow the patterns our solution extracts but are somewhat unclassifiable to human eyes. These samples are generated by assigning random values to the high energy parts of the extracted patterns. Although these samples do not look like the corresponding digits, in fact a good number of them do not look like digits, these samples are still classified into corresponding classes with confidences close to $100\%$. To some extent, this verifies that the extracted patterns provided by our solution are indeed important to DNN in this case.

We also extend our evaluation of scrutability to the scenario of text classification using `comp.sys' data. Table~\ref{tab:rule_text} lists two sets of keywords that our approach extracts for two ML models. Figure~\ref{fig:adv_text} shows one classification example, in which both learning models classify the snippet into `ibm.pc.hardware' with high confidence (94.20\% and 93.47\% for random forest and SVM, respectively). 

We replace word `{\em dos}' -- important for both classifiers -- with the words shown in Figure~\ref{fig:adv_text}, and test each of the newly crafted snippets against both classifiers. The value shown in Figure~\ref{fig:adv_text} indicates the confidences of categorizing new snippets into `ibm.pc.hardware'. We notice that by replacing `{\em dos}' with words that our solution deems important for another class (e.g. `mac.hardware'), we dramatically reduce the ML model's confidence in classifying the snippet under investigation as 'ibm.pc.hardware'. This again verifies the patterns that our approach extracts accurately reflect what are learned by both ML classifiers. 
\vspace{-2mm}
\section{Conclusion and Future Work}
\label{sec:conclusion}
\vspace{-2mm}

This work introduces a new technical approach to interrogate complicated ML models. Technically, it treats a target learning model as a black box and approximates its decision boundary through a Bayesian regression mixture model with multiple elastic nets. With this approach, model developers and users can approximate complex ML models with low errors and obtain better explanations of individual decisions. More importantly, they can extract patterns learned by a target learning model and use it to scrutinize model strengths and weaknesses. 

While our proposed approach exhibits outstanding performance in explaining individual decisions, and provides a user with an ability to discover model weaknesses, its performance may not be as good as what we observe in explaining discriminative learning model, particularly when applied to interrogating temporal learning models (e.g., hidden Markov models or recurrent neural networks). This is due to the fact that, our approach takes features independently whereas time series analysis deals with features temporally dependent. As part of the future work, we will therefore equip our approach with the ability of dissecting temporal learning models.

\small
\bibliography{ref}
\bibliographystyle{abbrv}


\end{document}